\documentclass[sigconf]{acmart}
\AtBeginDocument{%
  \providecommand\BibTeX{{%
    \normalfont B\kern-0.5em{\scshape i\kern-0.25em b}\kern-0.8em\TeX}}}

\usepackage{amsmath, amsfonts}
\usepackage{booktabs, multirow}
\usepackage{subfigure}
\usepackage{array,graphicx, adjustbox}
\usepackage{algorithm}
\usepackage{algorithmic}

\copyrightyear{2022} 
\acmYear{2022} 
\setcopyright{acmlicensed}\acmConference[SIGIR '22]{Proceedings of the 45th
International ACM SIGIR Conference on Research and Development in
Information Retrieval}{July 11--15, 2022}{Madrid, Spain}
\acmBooktitle{Proceedings of the 45th International ACM SIGIR Conference on
Research and Development in Information Retrieval (SIGIR '22), July 11--15,
2022, Madrid, Spain}
\acmPrice{15.00}
\acmDOI{10.1145/3477495.3531796}
\acmISBN{978-1-4503-8732-3/22/07}

\begin{document}

\fancyhead{}
\title{Value Penalized Q-Learning for Recommender Systems}

\author{Chengqian Gao}\authornote{Part of this work is done when taking an intership in Tencent AI Lab, Shenzhen.}
\affiliation{%
  \institution{Tsinghua University}
  \city{Shenzhen}
  \country{China}}
\email{gcq19@mails.tsinghua.edu.cn}

\author{Ke Xu}
\affiliation{%
  \institution{Tencent AI Lab}
  \city{Shenzhen}
  \country{China}}
\email{kaylakxu@tencent.com}

\author{Kuangqi Zhou}
\affiliation{%
  \institution{National University of Singapore}
  \country{Singapore}}
\email{kzhou@u.nus.edu}

\author{Lanqing Li}
\affiliation{%
 \institution{Tencent AI Lab}
 \city{Shenzhen}
 \country{China}}
\email{lanqingli1993@gmail.com}

\author{Xueqian Wang}
\affiliation{%
  \institution{Tsinghua University}
  \city{Shenzhen}
  \country{China}}
\email{wang.xq@sz.tsinghua.edu.cn}

\author{Bo Yuan}
\affiliation{%
  \institution{Tsinghua University}
  \city{Shenzhen}
  \country{China}}
\email{boyuan@ieee.org}

\author{Peilin Zhao}\authornote{Corresponding author.} 
\affiliation{%
  \institution{Tencent AI Lab}
  \city{Shenzhen}
  \country{China}}
\email{peilinzhao@hotmail.com}


\begin{abstract}
Scaling reinforcement learning (RL) to recommender systems (RS) is promising since maximizing the expected cumulative rewards for RL agents meets the objective of RS, i.e., improving customers' long-term satisfaction. 
A key approach to this goal is offline RL, which aims to learn policies from logged data rather than expensive online interactions. 
In this paper, we propose \textit{Value Penalized Q-learning (VPQ)}, a novel uncertainty-based offline RL algorithm that penalizes the unstable Q-values in the regression target using uncertainty-aware weights, achieving the conservative Q-function without the need of estimating the behavior policy, suitable for RS with a large number of items. 
Experiments on two real-world datasets show the proposed method serves as a \textit{gain plug-in} for existing RS models. 
\end{abstract}



\maketitle
\section{Introduction}



Practical recommender systems (RS) are usually trained to generate relevant items for users, considering less on their long-term utilities. 
Reinforcement learning (RL) methods maximize the discounted cumulative rewards over time, meeting the original needs of RS. 
Driven by this, there has been tremendous progress in developing reinforcement learning-based recommender systems (RLRS) \cite{afsar2021reinforcement}.


Directly utilizing RL algorithms to train agents from the offline data often results in poor performance \cite{lazic2018data, zhang2020cost, ye2020mastering}, even for off-policy methods, which can leverage data collected by other policies in principle \cite{fujimoto2019off, kumar2019stabilizing}. The main reason for such failures is the overestimation of out-of-distribution (OOD) queries \cite{levine2020offline}, i.e., the learned value function is trained on a small set of action space while is evaluated on all valid actions. 
Offline RL \cite{levine2020offline, fujimoto2019off} focuses on the problem of training agents from static datasets, addressing the OOD overestimation problem without interacting with the environment, making it achievable to build RLRS agents from offline datasets. 


There are still challenges for offline RL-based RS models due to the different settings between RL and RS tasks. 
The first is the large action space (the number of candidate items for RS agents is usually above 10k), which exacerbates the OOD overestimation problem and also makes it difficult to estimate the behavior policy that generated the datasets \cite{fujimoto2019off, kumar2019stabilizing, jaques2019way, wu2019behavior, wang2020off, kostrikov2021offline, Fujimoto2021AMA}. 
Another problem is the mismatch between relevant and valuable recommendations. RL approaches produce recommendations with the highest long-term utility while ignoring their relevance to users. 
Finally, the non-stationary dynamic of RS also differs from RL environments. In general offline RL problems, e.g., D4RL \cite{justinfu2020D4RL}, trajectories are collected by policies with an environment whose dynamics are exactly as same as the test environment. While in RS scenarios, the dynamics constantly change as the user preferences change over time. 

In this work, we propose Value Penalized Q-learning (VPQ), an uncertainty-based offline RL method, to alleviate the OOD overestimation issue. Then we integrate it into classic RS models, enhancing their capacity to generate relevant and valuable recommendations.

Specifically, we use a sequential RS model to represent the sequence of user-item interactions (mapping the last 10 items into hidden state $s_t$). 
The value function regresses on the discounted cumulative reward $Q(s_t,a_t) = \sum_{t=0}^{\inf} \gamma^t r(s_t,a_t)$ for each recommendation item $a_t$ with respect to the sate $s_t$, immediate reward $r$ (1.0 for purchases and 0.2 for clicks), and a discount factor $\gamma$.

Algorithmically, we take two techniques to address the OOD overestimation problem. 
Firstly, we stabilize the Q-predictions using an ensemble of Q-functions with different random learning rates and set the variance of their predictions as the uncertainty quantifier. 
Secondly, we penalize the unstable Q-values in regression target with uncertainty-aware weights, reducing the overestimation for OOD actions. The key component is to reduce the unstable predictions with the proposed $p-mul$ form penalty. We will detail it in the method section. 
In order to exploit the learned value function for the non-stationary environment, we employ the critic framework \cite{xin2020self}, to generate recommendations with both relevance and long-term satisfaction. 
To summarize: 
\begin{itemize}
    \item [(1)]
    We propose an uncertainty-based offline RL method, VPQ, to alleviate the OOD overestimation problem during training by penalizing the unstable predictions with uncertainty quantification, without estimating the behavior policy. 
    \item [(2)]
    We empirically show that it is more effective to attain the conservative value function by the proposed $p-mul$ form, i.e., multiplying the unstable predictions with uncertainty-aware weights, in the recommendation scenarios.
    \item [(3)]
    Extensive experiments show that the benefit of integrating with VPQ contains better representation and reinforcing actions with long-term rewards. 
\end{itemize}

\section{Methods}
We propose Value Penalized Q-learning (VPQ), an uncertainty-based offline RL algorithm, to estimate the long-term rewards for recommendation candidates. We then integrate it into classic RS models to generate relevant and valuable recommendations.

\subsection{Value Penalized Q-Learning}

\subsubsection{Preliminary: Q-Learning Algorithm.}
The Q-learning framework for recommendation has been adopted in many previous work \cite{zhao2018deep, zhao2018recommendations, zheng2018drn, zhao2019deep, xin2020self}. It usually learns a state-action value function $Q_{\theta}(s_t, a_t)$ by minimizing $\mathcal{L}_{\theta} =  \frac{1}{2} \mathop{\mathbb{E}}\limits_{(s_t,a_t,r_t,s_{t+1}) \sim D} \Big[ \big(  y - Q_{\theta}(s_t, a_t) \big)^2 \Big]$ with a dataset $D$, and a target $y = r + \gamma \max_{a}Q_{\theta_T}(s_{t+1}, a)$ with frozen parameters from historical value function $Q_{\theta}$.

However, when learning from static datasets, the $\max$ operator inhibits the performance as it queries all successor Q-values while only a small set of actions has been recorded. 
Unstable predictions on unseen actions undermine the learned Q-function and thus, in turn, exacerbate instability, resulting in erroneous Q-function when learning from static datasets \cite{kumar2019stabilizing, fujimoto2019off}. 

\subsubsection{Two Different Ways of Using the Uncertainty Metric.} An intuitive way to uncertainty-based offline RL \cite{levine2020offline, buckman2020importance, jin2020pessimism}, denoted as \textit{p-sub}, is removing the successor uncertainty from each query via:
\begin{equation}
    y = r + \gamma \max_a \big (Q(s_{t+1}, a) - \lambda \ {\rm Unc}(s_{t+1}, a) \big),
    \label{intuition_form}
\end{equation}
where Unc quantifies the amount of uncertainty for each Q query.

Our approach to penalizing the unstable prediction is: 
\begin{equation}
    y = r + \gamma \max_a \big (Q(s_{t+1}, a) \cdot W(s_{t+1}, a) \big),
    \label{VPQ_form}
\end{equation}
with an uncertainty-aware weight $W(s_{t+1},a)$ designed as:
\begin{equation}
      W(s_{t+1}, a) = \frac{1}{1 + \lambda \ {\rm Unc}(s_{t+1}, a)},
      \label{VPQ_example}
\end{equation}
with $\lambda>0$ controlling the strength of penalty on uncertain queries. ${\rm Unc}(s_{t+1},a)$ is defined as the standard deviation across the target Q-function ensemble, i.e., $\tilde \sigma_T(s_{t+1}, a) = SD \big ( \{ Q_{\theta_k}(s_{t+1}, a) \}_{k=1}^K\big)$. We denote this form as \textit{p-mul} for the stable predictions are estimated by multiplying the unstable values by uncertainty-aware weights.

\subsubsection{Analysis: Penalizing in a Stable Manner.}

Assuming that unstable predictions on OOD actions for a given state are i.i.d. random variables follow a normal distribution $x_i = Q(s_t,a_i) \sim \mathcal{N}(\mu,\sigma^2)$, then the target value for OOD actions (denoted as $y_{OOD}$) follows:
\begin{equation}
    y_{OOD} = r + \gamma \max_{1\leq i \leq n; x_i \sim \mathcal{N}(\mu, \sigma^2)} x_i,
\end{equation}
where $n$ is close to the number of candidate items in RS, and for brevity, we denote $Q_T$ as the output of the $\max$ operation. 
We can approximate the expectation of $Q_T$ through the form \cite{blom1958, harter1961, royston1982expected}: 
\begin{equation}
    \mathbb{E}\Big[ \max_{1\leq i \leq n; x_i \sim \mathcal{N}(\mu, \sigma^2)} x_i\Big] \approx \mu + \sigma \Phi^{-1}(\frac{n - 0.375}{n - 2\times0.375 + 1}),
\end{equation}
where $\Phi^{-1}$ is the inverse of the standard normal CDF.

\begin{algorithm}[ht]
    \caption{VPQ: Value Penalized Q-Learning}
    \label{algo}

    \begin{algorithmic}[1]
    \REQUIRE   K $Q$-functions with parameters $\{\theta_k\}_{k=1}^K$,\\ a categorical distribution $P_{\Delta}$ for Random Ensemble Mixture,\\ a scale factor $\lambda$ for VPQ, and an offline dataset $D$. \\
    \ENSURE   $\{\theta_k\}_{k=1}^K$
    
    \STATE Initialize $\{\theta_k\}_{k=1}^K$.
    \WHILE{not done}

    \STATE Sample a mini-batch of transitions $(s_t, a_t, r, s_{t+1})$ randomly from $D$
    \STATE Sample mixture weights $\{ \alpha_k \}_{k=1}^K$ from $P_{\Delta}$
    \STATE Compute the sample standard deviation $\tilde \sigma_T(s_{t+1},a)$
    \STATE Compute random mixture of target networks:
    \\ \qquad  $\tilde \mu_T(s_{t+1}, a) = \sum_{k=1}^K \alpha_k Q_{\theta^T_k}(s_{t+1},a)$

    \STATE Compute the uncertainty-aware weight:
    \\ \qquad  $W(s_{t+1}, a)= \big(1 + \lambda \  \tilde \sigma_T(s_{t+1}, a)\big)^{-1}$
        
    \STATE Compute the penalized target:
    \\ \qquad  $y_t = r + \gamma \max_{a} \big( \tilde \mu_T(s_{t+1}, a) \cdot W(s_{t+1}, a)\big)$
    
    \STATE Update each Q-function with the mixture weight $\alpha_k$: 
    \\ \qquad  $\mathcal{L}_{\theta} =  \frac{1}{2}\big(y_t - \sum_{k=1}^K \alpha_k \cdot Q_{\theta_k}(s_t, a_t) \big)^2$
    \ENDWHILE
    \end{algorithmic}
\end{algorithm} 

By using the properties of the expectation and $\max$ operator, we have expectation of penalized $Q_T$ with the \textit{\textit{p-sub}} formulation:
\begin{equation}
    \mathbb{E}\Big[ Q_T - \lambda \sigma  \Big] = \mu + (C_0 - \lambda) \sigma \\
\label{approximate:sub}
\end{equation}
and the expected value of that with the \textit{p-mul} form:
\begin{equation}
    \mathbb{E} \Big[ \frac{Q_T}{1 + \lambda \sigma} \Big] = \frac{1}{1 + \lambda \sigma} (\mu + \sigma C_0),
\end{equation}
where $C_0$ is a constant number for a given $n$.

\begin{figure}[htb]
  \centering
  \subfigure[p-sub]{
  \label{Demo_minus}
  \centering
  \includegraphics[width=0.75\linewidth]{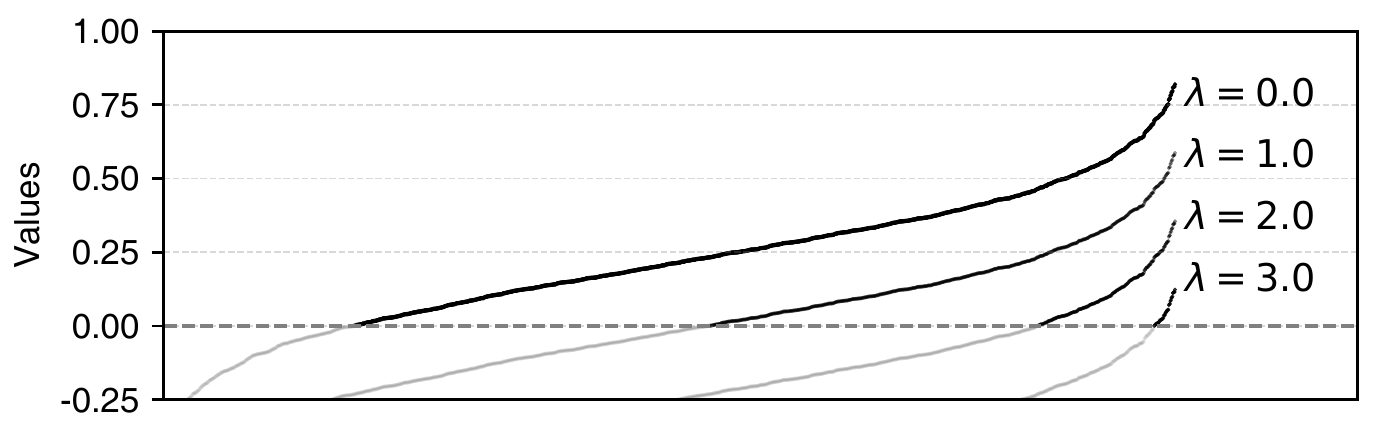}
  }\vspace{-0.2cm}
  \subfigure[p-mul]{
  \label{Demo_multiply}
  \centering
  \includegraphics[width=0.75\linewidth]{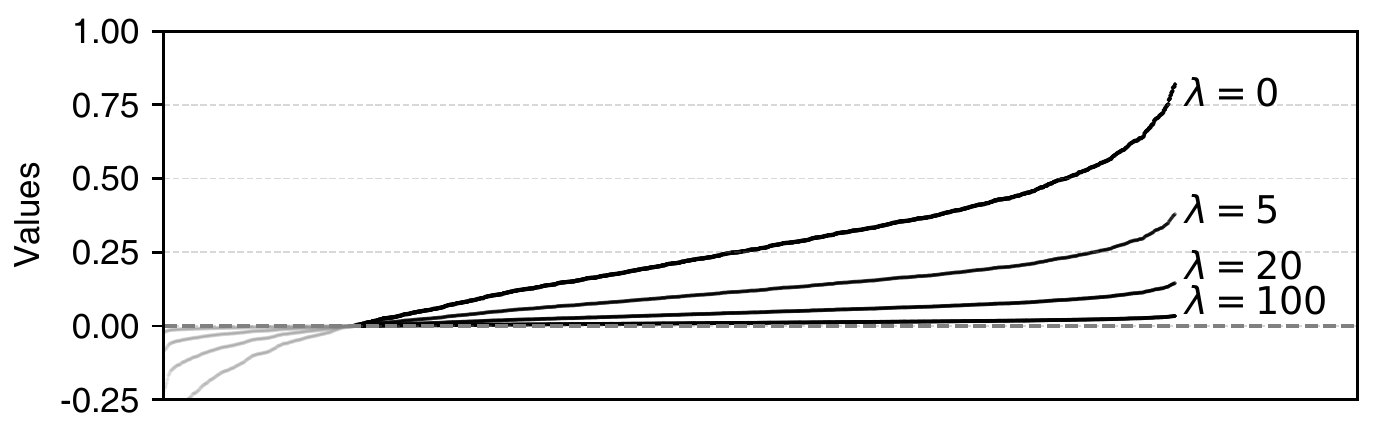}
  }\vspace{-0.3cm}
  \caption{
  Penalize the simulated unstable predictions in different ways. Heavy black lines are dense points sampled from a Gaussian distribution and sorted by their values. Scaling factor $\lambda$ controls the strength of penalization.
  }\vspace{-0.3cm}
  \label{suppress}
\end{figure}

The proposed form of penalization has two advantages. 
Firstly, the \textit{p-mul} form is more robust. 
For training RLRS agents, $C_0$ increases with the dimension of action space. To reduce unstable predictions, \textit{p-sub} has to increase its $\lambda$ in Equation (\ref{approximate:sub}), increasing the risk of producing negative Q-values, and thus resulting in unstable training. By contrast, the penalty form \textit{p-mul} would always keep the target value greater than zero, even with a large scale factor \footnote{In our setting, the Q-value should be a positive number, with reward $r=0, 0.2, 1.0$.}. 
Secondly, the proposed \textit{p-mul} can heavily penalize the unstable and large predictions, while the \textit{p-sub} formulation mainly concerns the small and unstable Q-values and thus may fail to obtain a conservative prediction. To illustrate this, we provide a toy experiment in Figure \ref{suppress}.

A keen reader may note that the proposed \textit{p-mul} form fails to penalize Q-values that less than zero. However, we argue that such failures can be avoided via a linear reward transformation \cite{ng1999policy}. 

\vspace{-0.2cm}

\subsubsection{Analysis: Uncertrainty-Aware Discount Factor.} We note the proposed uncertainty weight can be absorbed into the discount factor term $\gamma$ in target, i.e., $y= r + \max_a [\gamma \cdot W(s_{t+1}, a)] \cdot Q(s_{t+1}, a)$. This motivates the proposed VPQ algorithm in another perspective, as $\gamma$ affects the performance of the learned value function \cite{kumar2019stabilizing}:
\begin{align}
    \lim_{k \to \infty}\mathbb{E}_{d_0}\big[\Big|V^{\pi^k}(s_t)  &- V^\Pi(s_t)\Big|\big] \le \nonumber \\& \frac{2\gamma}{(1-\gamma)^2} C_{\Pi, \mu} \mathbb{E_{\mu}}\Big[ \max_{\pi \in \Pi} \mathbb{E}_\pi [\delta(s_t,a_t)]\Big],
    \label{inequality}
\end{align}
\vspace{-0.1cm}
with $\Pi$ for a set of policies, $V^{\Pi}$ for the fixed point that the constrained Bellman backup $\mathcal{T}^\Pi$ convergences to, concentrability coefficient $C_{\Pi, \mu}$ for quantifying how far the distribution of the policy action $\pi(a_t |s_t) \sim \Pi$ is from the corresponding dataset action $\mu(a_t |s_t)$, and $\delta(s_t,a_t) \ge \sup_k |Q^k(s_t,a_t) - \mathcal{T}^\Pi Q^{k-1}(s_t,a_t)|$ bounds the Bellman error.

The proposed weight $W$ affects the bound (\ref{inequality}) through the absorbed term $\frac{2 \gamma W}{(1-\gamma W)^2}$. For example, for two queries with discount factor $\gamma=0.99$ and uncertainty-aware weight $W = 0.9, 0.5$, the absorbed term can be 149.98 and 3.88. In this way, the proposed $p-mul$ form controls how much importance we assign to future rewards with uncertainty. 

\subsubsection{Details about the Proposed Algorithm.} 
Inspired by the powerful penalty form, \textit{p-mul}, we developed VPQ. 
It contains an ensemble of Q-functions with the Random Ensemble Mixture (REM) technique \cite{agarwal2020optimistic} for diversity and assigns the standard deviation of Q-values across the ensemble as an uncertainty indicator.

VPQ eliminates the need to estimate the difficult behavior policy for constraint \cite{fujimoto2019off, kumar2019stabilizing, jaques2019way, wu2019behavior, wang2020off, kostrikov2021offline, Fujimoto2021AMA}. Besides, as an uncertainty-based method, VPQ enjoys benefits from the progress in uncertainty measurement. Finally, we summarize our method in Algorithm \ref{algo}.

\subsection{Relevant v.s. Valuable} 
So far, we have developed VPQ to indicate the long-term rewards for recommendation candidates. However, there exists another concern for RLRS, i.e., how to exploit the learned agent. Generating recommendations via $a=\arg\max_a Q(s_t,a)$ leads to valuable recommendations without considering their relevance, while classic sequential RS models recall and/or rank relevant items ignoring their long-term utilities. To tackle this, we introduce the critic framework \cite{xin2020self}, which is inspired by \citet{konda2000actor}. 

Specifically, we use the classic sequential RS model to extract hidden states for the input interaction sequences and map them to two types of outputs, Q-values through the Q head and classification logits using a CE head. We optimize the Q head with VPQ algorithm and minimize the reweighted cross-entropy loss for CE head, i.e.,
\begin{equation}
    \mathcal{L}_{\phi} = CE(s_t,a) \cdot \ {\rm no\_gradient}(Q(s_t,a))
    \label{critic_framework}
\end{equation}
At test time, we only use the CE head to generate recommendations. 
Although it is a trade-off between the classic recommendation paradigms and alluring RL methods, the critic framework improves the performance of RS with stability, as shown in our experiments.

\begin{table*}[!htp]
  \caption{Overall performance comparison on two recommendation datasets, averaged over 5 runs. NG is short for NDCG.}
  \label{vs_baseline}
  \centering
  \begin{tabular}{lcccccccccc}
    \toprule
    \multirow{3}{*}{Models} 
    & \multicolumn{4}{c}{Retailrocket} & \multicolumn{4}{c}{Yoochoose} & \multirow{3}{*}{Total} \\ 
    \cmidrule(r){2-5}   \cmidrule(r){6-9} 
    
    ~ & \multicolumn{2}{c}{purchase} & \multicolumn{2}{c}{click} & \multicolumn{2}{c}{purchase} & \multicolumn{2}{c}{click} ~ \\
    \cmidrule(r){2-3}   \cmidrule(r){4-5} \cmidrule(r){6-7} \cmidrule(r){8-9} 
 
    ~ & HR@20 & NG@20 & HR@20 & NG@20 & HR@20 & NG@20 & HR@20 & NG@20 & ~\\


    \midrule
    Caser &
    .4226 & .3076 & .2813 & .1875 & .6607 & .3911 & .4613 & .2545 & 2.9665 (0.00\%)\\
    Caser-SAC &
    .4617 & .3336 & .3006 & .1985 & .6889 & .4173 & .4513 & .2472 & 3.0991 (4.47\%)\\
    Caser-REM & 
    .4711 & .3390 & .3019 & .1999 & .6916 & .4216 & .4481 & .2452 & 3.1184 (5.12\%)\\
    Caser-Minus & 
    .4618 & .3325 & .2991 & .1984 & .7029 & .4330 & .4429 & .2419 & 3.1124 (4.92\%)\\
    Caser-CQL &
    .4699 & .3376 & .3048 & .2009 & .6705 & .4008 & .4598 & .2527 & 3.0970 (4.40\%)\\
    Caser-UWAC & 
    .3884 & .2851 & .2588 & .1714 & .6758 & .4126 & .4509 & .2464 & 2.8893 (-2.60\%)\\
    Caser-VPQ & 
    \textbf{.4775} & \textbf{.3454} & \textbf{.3067} & \textbf{.2036} & \textbf{.7081} & \textbf{.4350} & .4459 & .2439 & \textbf{3.1661 (6.73\%)}\\
    
    \midrule
    Next & 
    .6564 & .4961 & .3329 & .2130 & .5934 & .3352 & .4937 & .2715 & 3.3922 (0.00\%)\\
    Next-SAC & 
    .6741 & .5354 & .3346 & .2149 & .5899 & .3383 & .4730 & .2584 & 3.4186 (0.78\%)\\
    Next-REM & 
    .6839 & .5387 & .3433 & .2216 & .6025 & .3509 & .4781 & .2623 & 3.4812 (2.62\%)\\
    Next-Minus & 
    .6835 & .5432 & .3446 & .2227 & .6047 & .3511 & .4771 & .2622 & 3.4890 (2.85\%)\\
    Next-CQL & 
    .6845 & .5399 & .3423 & .2215 & .6073 & .3509 & .4802 & .2635 & 3.4901 (2.89\%)\\
    Next-UWAC &
    .6840 & .5310 & .3452 & .2206 & .6155 & .3583 & .4861 & .2694 & 3.5102 (3.48\%)\\
    Next-VPQ & 
    \textbf{.6990} & \textbf{.5714} & \textbf{.3556} & \textbf{.2330} & \textbf{.6226} & \textbf{.3672} & .4944 & \textbf{.2744} & \textbf{3.6175 (6.64\%)}\\
    
    \midrule
    SASRec & 
    .6387 & .4599 & .3516 & .2190 & .6630 & .3733 & .5044 & .2761 & 3.4861 (.00\%)\\
    SASRec-SAC & 
    .6981 & .5629 & .3603 & .2347 & .6786 & .3955 & .5015 & .2760 & 3.7077 (6.35\%)\\
    SASRec-REM & 
    .6655 & .5113 & .3650 & .2361 & .6809 & .3986 & .5033 & .2779 & 3.6387 (4.37\%)\\
    SASRec-Minus & 
    .6666 & .5125 & .3634 & .2353 & .6815 & .3965 & .5038 & .2783 & 3.6378 (4.35\%)\\
    SASRec-CQL & 
    .7046 & .5599 & .3749 & .2404 & .6631 & .3805 & .5013 & .2751 & 3.6997 (6.12\%)\\
    SASRec-UWAC & 
    .6657 & .5015 & .3715 & .2367 & .6786 & .3959 & .5126 & \textbf{.2850} & 3.6475 (4.63\%)\\
    SASRec-VPQ & 
    \textbf{.7171} & \textbf{.5914} & \textbf{.3785} & \textbf{.2484} & \textbf{.6841} & \textbf{.4063} & .5081 & .2814 & \textbf{3.8153 (9.44\%)}\\
  \bottomrule
\end{tabular}
\end{table*}
\begin{figure*}[ht]
    \subfigure[Retailrocket]{
    \label{RQ3_Retailrocket}
    \includegraphics[width=.47\linewidth]{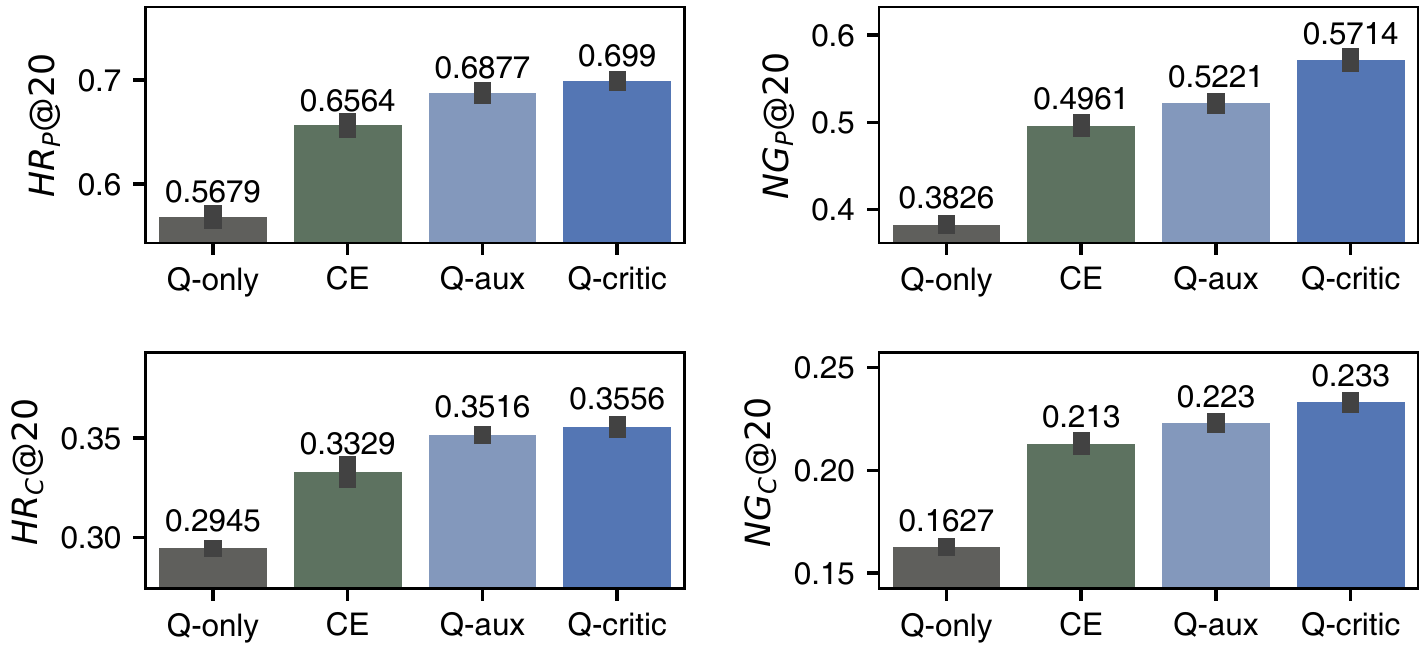}
    }
    \subfigure[Yoochoose]{
    \label{RQ3_Yoochoose}
    \includegraphics[width=.47\linewidth]{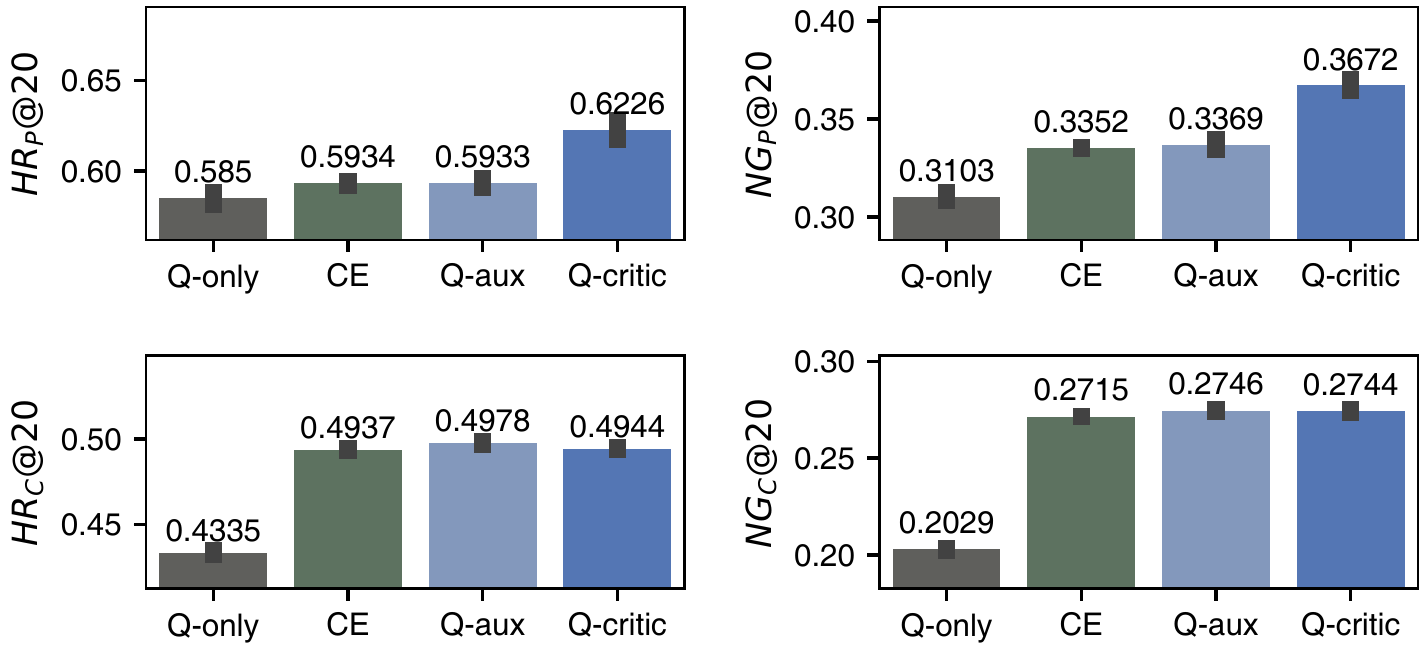}
    }\vspace{-0.1cm}
    \caption{Comparison of performance of utilizing the learned Q-function in different ways. Error bars show standard deviations.}
    \label{RQ3_performance}
\end{figure*}

\begin{figure}[!htp]
    \centering
    \includegraphics[width=0.9\linewidth]{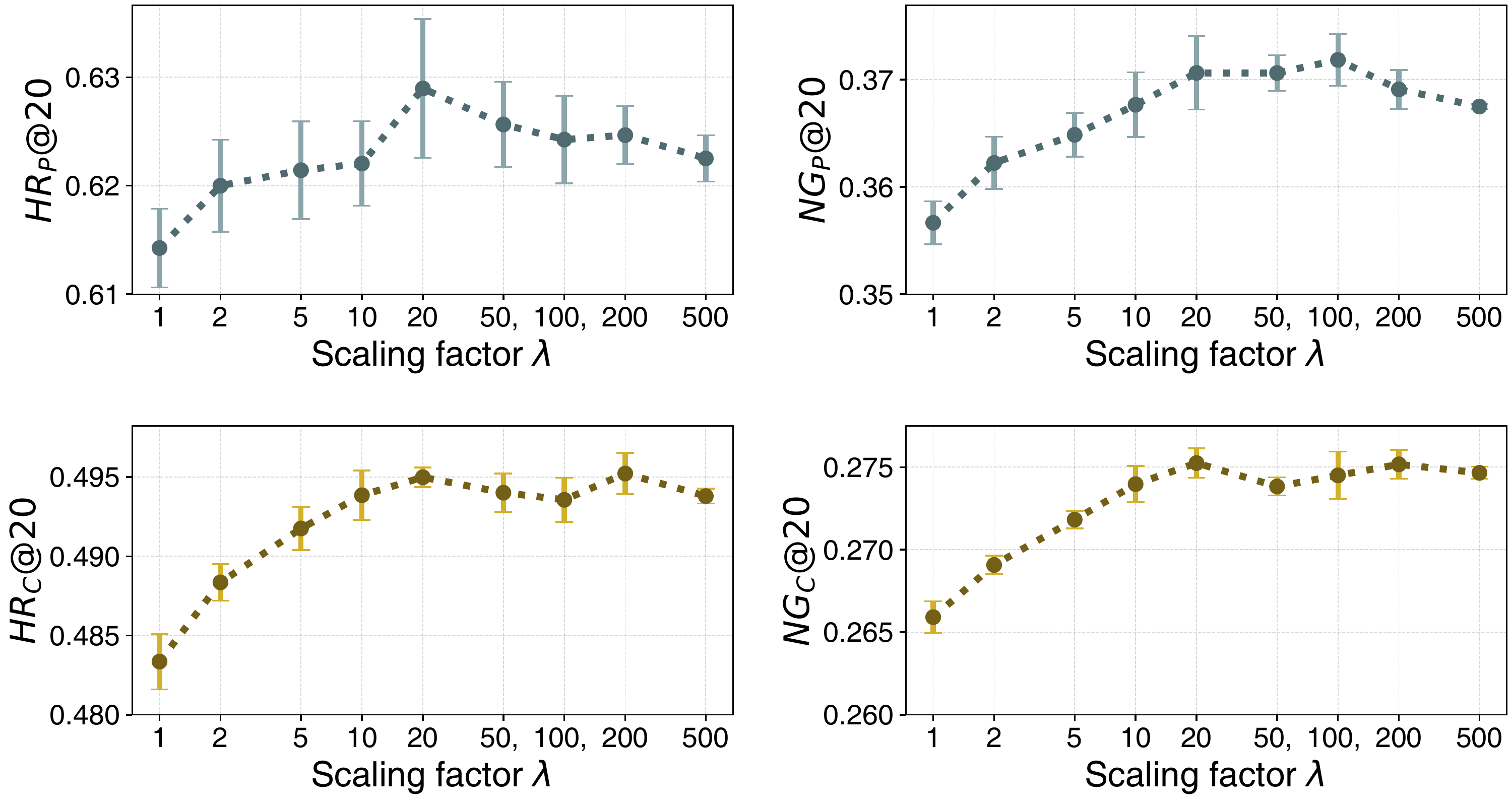}
    \vspace{-0.2cm}
    \caption{Sweeping the scaling factor on Yoochoose dataset.}
    \label{RQ4_parapmeter_search}
\end{figure}

\section{Experiments}
In order to verify the effectiveness of the proposed VPQ, we conduct experiments \footnote{Code is available at https://drive.google.com/drive/folders/1i3E1QTkscyoCAXRH\\28OxNLpbJeH19sSm?usp=sharing} on two standard recommender systems datasets, namely Retailrocket \footnote{https://www.kaggle.com/retailrocket/ecommerce-dataset} and Yoochoose \footnote{https://recsys.acm.org/recsys15/challenge/}, following the data processing details in \citet{xin2020self}. 
We use two metrics: HR (hit ratio) for recall and NDCG (normalized discounted cumulative gain) for rank and compute the performance on clicked and ordered items separately. For example, we define HR for clicks as: 
\begin{equation}
    \rm HR(click) = \frac{\# \ hits \  among \ clicks}{\#  \ clicks}
\end{equation}

\subsection{Performance Gains from VPQ}
We integrate VPQ with three classic sequential RS methods: Caser \cite{tang2018personalized}, NextItNet \cite{yuan2019simple} and SASRec \cite{kang2018self}. We also compare its performance with the following online and offline RL-based methods: {SAC} \cite{xin2020self}, {REM} \cite{agarwal2020optimistic}, {Minus} (involved the \textit{p-sub} formulation in Equation (\ref{intuition_form})), {CQL} \cite{kumar2020conservative}, and {UWAC} \cite{yuewu2021}. 
All the above methods are embedded with the classic RS models under the critic framework. We select the best hyper-parameter $\lambda$ for Minus, $\alpha$ for CQL, and $\beta$ for UWAC by sweeping on each dataset. $\lambda$ for VPQ is set to 20.

The consistent performance improvement shown in Table \ref{vs_baseline} demonstrates that the integration with the learned Q-function generally outperforms the original RS models. 
Compared to other baseline algorithms, the proposed VPQ achieves more performance gains, illustrating the effectiveness of our approach.

\subsection{Ablation Study}
We conduct ablation study with four settings: (1) \textbf{Q-only}, which generates recommendations with the highest Q-values (the most valuables), (2) \textbf{CE}, which generates recommendations with only the original model (the most relevant ones), (3) \textbf{Q-aux}, minimizing the TD-error as an auxiliary task without utilizing the Q-values (for representation), (4) \textbf{Q-critic}, reinforcing the recommendation actions with Q-values (using the loss function in Equation (\ref{critic_framework})). 

Results are shown in Figure \ref{RQ3_performance}. 
\textbf{Q-only} achieves the worst performance on both datasets, illustrating that valuable recommendations (items with high Q-values) tend to lose relevance. 
\textbf{Q-aux} surpasses \textbf{CE}, suggesting that minimizing the TD-error benefits representation learning. 
The results above suggest that improvement of \textbf{Q-critic} comes from better state representation (auxiliary loss) and exploiting accurate Q-values (reinforcing the valuable actions). 

\subsection{Hyper-parameter Study}
The scaling factor $\lambda$ controls the strength of uncertainty penalty for VPQ (Equation \ref{VPQ_example}) and thereby affects the performance. For a small scaling factor, VPQ degrades to REM \cite{agarwal2020optimistic} without explicit penalty on OOD predictions. With a large one, VPQ tends to regress on immediate rewards rather than the long-term rewards. Results in Figure \ref{RQ4_parapmeter_search} verifies this on the Next-VPQ model. The best $\lambda$ for the other two base models and the Retailrocket dataset is also 20. 

\section{Conclusion}
Directly scaling RL methods into offline setting often end up with unsatisfied results. This work proposes an uncertainty-based offline RL method, \textit{Value Penalized Q-Learning (VPQ)}, which captures uncertainty information across an ensemble of Q-functions to tackle the overestimations on OOD actions and thus estimates the long-term rewards for RLRS, without estimating the diffcult behavior policy and thus suitable for RS scenarios. Evaluations on two standard RS datasets demonstrate that the proposed methods could serve as a gain plug-in for classic sequential RS models.

\bibliographystyle{ACM-Reference-Format}
\bibliography{my_refs}


\begin{thebibliography}{31}


\ifx \showCODEN    \undefined \def \showCODEN     #1{\unskip}     \fi
\ifx \showDOI      \undefined \def \showDOI       #1{#1}\fi
\ifx \showISBNx    \undefined \def \showISBNx     #1{\unskip}     \fi
\ifx \showISBNxiii \undefined \def \showISBNxiii  #1{\unskip}     \fi
\ifx \showISSN     \undefined \def \showISSN      #1{\unskip}     \fi
\ifx \showLCCN     \undefined \def \showLCCN      #1{\unskip}     \fi
\ifx \shownote     \undefined \def \shownote      #1{#1}          \fi
\ifx \showarticletitle \undefined \def \showarticletitle #1{#1}   \fi
\ifx \showURL      \undefined \def \showURL       {\relax}        \fi
\providecommand\bibfield[2]{#2}
\providecommand\bibinfo[2]{#2}
\providecommand\natexlab[1]{#1}
\providecommand\showeprint[2][]{arXiv:#2}

\bibitem[Afsar et~al\mbox{.}(2021)]%
        {afsar2021reinforcement}
\bibfield{author}{\bibinfo{person}{M~Mehdi Afsar}, \bibinfo{person}{Trafford
  Crump}, {and} \bibinfo{person}{Behrouz Far}.}
  \bibinfo{year}{2021}\natexlab{}.
\newblock \bibinfo{title}{Reinforcement learning based recommender systems: A
  survey}.
\newblock
\newblock
\showeprint[arxiv]{2101.06286}


\bibitem[Agarwal et~al\mbox{.}(2020)]%
        {agarwal2020optimistic}
\bibfield{author}{\bibinfo{person}{Rishabh Agarwal}, \bibinfo{person}{Dale
  Schuurmans}, {and} \bibinfo{person}{Mohammad Norouzi}.}
  \bibinfo{year}{2020}\natexlab{}.
\newblock \showarticletitle{An Optimistic Perspective on Offline Reinforcement
  Learning}. In \bibinfo{booktitle}{\emph{{ICML} 2020, 13-18 July 2020, Virtual
  Event}} \emph{(\bibinfo{series}{Proceedings of Machine Learning Research},
  Vol.~\bibinfo{volume}{119})}. \bibinfo{publisher}{{PMLR}},
  \bibinfo{pages}{104--114}.
\newblock
\urldef\tempurl%
\url{http://proceedings.mlr.press/v119/agarwal20c.html}
\showURL{%
\tempurl}


\bibitem[Blom(1958)]%
        {blom1958}
\bibfield{author}{\bibinfo{person}{Gunnar Blom}.}
  \bibinfo{year}{1958}\natexlab{}.
\newblock \bibinfo{booktitle}{\emph{Statistical Estimates and Transformed Beta
  Variables.}}
\newblock \bibinfo{publisher}{Almqvist \& Wiksell, John Wiley \& Sons, Inc.},
  \bibinfo{address}{Sweden}.
\newblock


\bibitem[Buckman et~al\mbox{.}(2020)]%
        {buckman2020importance}
\bibfield{author}{\bibinfo{person}{Jacob Buckman}, \bibinfo{person}{Carles
  Gelada}, {and} \bibinfo{person}{Marc~G. Bellemare}.}
  \bibinfo{year}{2020}\natexlab{}.
\newblock \bibinfo{title}{The Importance of Pessimism in Fixed-Dataset Policy
  Optimization}.
\newblock
\newblock
\showeprint[arxiv]{2009.06799}
\urldef\tempurl%
\url{https://arxiv.org/abs/2009.06799}
\showURL{%
\tempurl}


\bibitem[Fu et~al\mbox{.}(2020)]%
        {justinfu2020D4RL}
\bibfield{author}{\bibinfo{person}{Justin Fu}, \bibinfo{person}{Aviral Kumar},
  \bibinfo{person}{Ofir Nachum}, \bibinfo{person}{George Tucker}, {and}
  \bibinfo{person}{Sergey Levine}.} \bibinfo{year}{2020}\natexlab{}.
\newblock \bibinfo{title}{{D4RL:} Datasets for Deep Data-Driven Reinforcement
  Learning}.
\newblock
\newblock
\showeprint[arxiv]{2004.07219}
\urldef\tempurl%
\url{https://arxiv.org/abs/2004.07219}
\showURL{%
\tempurl}


\bibitem[Fujimoto and Gu(2021)]%
        {Fujimoto2021AMA}
\bibfield{author}{\bibinfo{person}{Scott Fujimoto} {and}
  \bibinfo{person}{Shixiang~Shane Gu}.} \bibinfo{year}{2021}\natexlab{}.
\newblock \bibinfo{title}{A Minimalist Approach to Offline Reinforcement
  Learning}.
\newblock
\newblock
\showeprint[arxiv]{2106.06860}
\urldef\tempurl%
\url{https://arxiv.org/abs/2106.06860}
\showURL{%
\tempurl}


\bibitem[Fujimoto et~al\mbox{.}(2019)]%
        {fujimoto2019off}
\bibfield{author}{\bibinfo{person}{Scott Fujimoto}, \bibinfo{person}{David
  Meger}, {and} \bibinfo{person}{Doina Precup}.}
  \bibinfo{year}{2019}\natexlab{}.
\newblock \showarticletitle{Off-Policy Deep Reinforcement Learning without
  Exploration}. In \bibinfo{booktitle}{\emph{{ICML} 2019, 9-15 June 2019, Long
  Beach, California, {USA}}} \emph{(\bibinfo{series}{Proceedings of Machine
  Learning Research}, Vol.~\bibinfo{volume}{97})}. \bibinfo{publisher}{{PMLR}},
  \bibinfo{pages}{2052--2062}.
\newblock
\urldef\tempurl%
\url{http://proceedings.mlr.press/v97/fujimoto19a.html}
\showURL{%
\tempurl}


\bibitem[Harter(1961)]%
        {harter1961}
\bibfield{author}{\bibinfo{person}{H.~Leon Harter}.}
  \bibinfo{year}{1961}\natexlab{}.
\newblock \showarticletitle{Expected values of normal order statistics}.
\newblock \bibinfo{journal}{\emph{Biometrika}} \bibinfo{volume}{48},
  \bibinfo{number}{1 and 2} (\bibinfo{year}{1961}), \bibinfo{pages}{151--165}.
\newblock


\bibitem[Jaques et~al\mbox{.}(2019)]%
        {jaques2019way}
\bibfield{author}{\bibinfo{person}{Natasha Jaques}, \bibinfo{person}{Asma
  Ghandeharioun}, \bibinfo{person}{Judy~Hanwen Shen}, \bibinfo{person}{Craig
  Ferguson}, \bibinfo{person}{{\`{A}}gata Lapedriza}, \bibinfo{person}{Noah
  Jones}, \bibinfo{person}{Shixiang Gu}, {and} \bibinfo{person}{Rosalind~W.
  Picard}.} \bibinfo{year}{2019}\natexlab{}.
\newblock \bibinfo{title}{Way Off-Policy Batch Deep Reinforcement Learning of
  Implicit Human Preferences in Dialog}.
\newblock
\newblock
\showeprint[arxiv]{1907.00456}
\urldef\tempurl%
\url{http://arxiv.org/abs/1907.00456}
\showURL{%
\tempurl}


\bibitem[Jin et~al\mbox{.}(2020)]%
        {jin2020pessimism}
\bibfield{author}{\bibinfo{person}{Ying Jin}, \bibinfo{person}{Zhuoran Yang},
  {and} \bibinfo{person}{Zhaoran Wang}.} \bibinfo{year}{2020}\natexlab{}.
\newblock \bibinfo{title}{Is Pessimism Provably Efficient for Offline RL?}
\newblock
\newblock
\showeprint[arxiv]{2012.15085}
\urldef\tempurl%
\url{https://arxiv.org/abs/2012.15085}
\showURL{%
\tempurl}


\bibitem[Kang and McAuley(2018)]%
        {kang2018self}
\bibfield{author}{\bibinfo{person}{Wangcheng Kang} {and}
  \bibinfo{person}{Julian~J. McAuley}.} \bibinfo{year}{2018}\natexlab{}.
\newblock \showarticletitle{Self-Attentive Sequential Recommendation}. In
  \bibinfo{booktitle}{\emph{{ICDM} 2018, Singapore, November 17-20, 2018}}.
  \bibinfo{publisher}{{IEEE} Computer Society}, \bibinfo{pages}{197--206}.
\newblock
\urldef\tempurl%
\url{https://doi.org/10.1109/ICDM.2018.00035}
\showDOI{\tempurl}


\bibitem[Konda(2002)]%
        {konda2000actor}
\bibfield{author}{\bibinfo{person}{Vijaymohan Konda}.}
  \bibinfo{year}{2002}\natexlab{}.
\newblock \emph{\bibinfo{title}{Actor-critic Algorithms}}.
\newblock \bibinfo{thesistype}{Ph.\,D. Dissertation}.
  \bibinfo{school}{Massachusetts Institute of Technology, Cambridge, MA,
  {USA}}.
\newblock
\urldef\tempurl%
\url{http://hdl.handle.net/1721.1/8120}
\showURL{%
\tempurl}


\bibitem[Kostrikov et~al\mbox{.}(2021)]%
        {kostrikov2021offline}
\bibfield{author}{\bibinfo{person}{Ilya Kostrikov}, \bibinfo{person}{Rob
  Fergus}, \bibinfo{person}{Jonathan Tompson}, {and} \bibinfo{person}{Ofir
  Nachum}.} \bibinfo{year}{2021}\natexlab{}.
\newblock \showarticletitle{Offline Reinforcement Learning with Fisher
  Divergence Critic Regularization}. In \bibinfo{booktitle}{\emph{{ICML} 2021,
  18-24 July 2021, Virtual Event}} \emph{(\bibinfo{series}{Proceedings of
  Machine Learning Research}, Vol.~\bibinfo{volume}{139})}.
  \bibinfo{publisher}{{PMLR}}, \bibinfo{pages}{5774--5783}.
\newblock
\urldef\tempurl%
\url{http://proceedings.mlr.press/v139/kostrikov21a.html}
\showURL{%
\tempurl}


\bibitem[Kumar et~al\mbox{.}(2019)]%
        {kumar2019stabilizing}
\bibfield{author}{\bibinfo{person}{Aviral Kumar}, \bibinfo{person}{Justin Fu},
  \bibinfo{person}{Matthew Soh}, \bibinfo{person}{George Tucker}, {and}
  \bibinfo{person}{Sergey Levine}.} \bibinfo{year}{2019}\natexlab{}.
\newblock \showarticletitle{Stabilizing Off-Policy Q-Learning via Bootstrapping
  Error Reduction}. In \bibinfo{booktitle}{\emph{{NeurIPS} 2019, December 8-14,
  2019, Vancouver, BC, Canada}}. \bibinfo{pages}{11761--11771}.
\newblock
\urldef\tempurl%
\url{https://proceedings.neurips.cc/paper/2019/hash/c2073ffa77b5357a498057413bb09d3a-Abstract.html}
\showURL{%
\tempurl}


\bibitem[Kumar et~al\mbox{.}(2020)]%
        {kumar2020conservative}
\bibfield{author}{\bibinfo{person}{Aviral Kumar}, \bibinfo{person}{Aurick
  Zhou}, \bibinfo{person}{George Tucker}, {and} \bibinfo{person}{Sergey
  Levine}.} \bibinfo{year}{2020}\natexlab{}.
\newblock \showarticletitle{Conservative Q-Learning for Offline Reinforcement
  Learning}. In \bibinfo{booktitle}{\emph{{NeurIPS} 2020, December 6-12, 2020,
  virtual}}.
\newblock
\urldef\tempurl%
\url{https://proceedings.neurips.cc/paper/2020/hash/0d2b2061826a5df3221116a5085a6052-Abstract.html}
\showURL{%
\tempurl}


\bibitem[Lazic et~al\mbox{.}(2018)]%
        {lazic2018data}
\bibfield{author}{\bibinfo{person}{Nevena Lazic}, \bibinfo{person}{Craig
  Boutilier}, \bibinfo{person}{Tyler Lu}, \bibinfo{person}{Eehern Wong},
  \bibinfo{person}{Binz Roy}, \bibinfo{person}{M.~K. Ryu}, {and}
  \bibinfo{person}{Greg Imwalle}.} \bibinfo{year}{2018}\natexlab{}.
\newblock \showarticletitle{Data center cooling using model-predictive
  control}. In \bibinfo{booktitle}{\emph{{NeurIPS} 2018, December 3-8, 2018,
  Montr{\'{e}}al, Canada}}. \bibinfo{pages}{3818--3827}.
\newblock
\urldef\tempurl%
\url{https://proceedings.neurips.cc/paper/2018/hash/059fdcd96baeb75112f09fa1dcc740cc-Abstract.html}
\showURL{%
\tempurl}


\bibitem[Levine et~al\mbox{.}(2020)]%
        {levine2020offline}
\bibfield{author}{\bibinfo{person}{Sergey Levine}, \bibinfo{person}{Aviral
  Kumar}, \bibinfo{person}{George Tucker}, {and} \bibinfo{person}{Justin Fu}.}
  \bibinfo{year}{2020}\natexlab{}.
\newblock \bibinfo{title}{Offline Reinforcement Learning: Tutorial, Review, and
  Perspectives on Open Problems}.
\newblock
\newblock
\showeprint[arxiv]{2005.01643}
\urldef\tempurl%
\url{https://arxiv.org/abs/2005.01643}
\showURL{%
\tempurl}


\bibitem[Ng et~al\mbox{.}(1999)]%
        {ng1999policy}
\bibfield{author}{\bibinfo{person}{Andrew~Y. Ng}, \bibinfo{person}{Daishi
  Harada}, {and} \bibinfo{person}{Stuart~J. Russell}.}
  \bibinfo{year}{1999}\natexlab{}.
\newblock \showarticletitle{Policy Invariance Under Reward Transformations:
  Theory and Application to Reward Shaping}. In
  \bibinfo{booktitle}{\emph{{(ICML} 1999), Bled, Slovenia, June 27 - 30,
  1999}}. \bibinfo{publisher}{Morgan Kaufmann}, \bibinfo{pages}{278--287}.
\newblock


\bibitem[Royston(1982)]%
        {royston1982expected}
\bibfield{author}{\bibinfo{person}{J.~P. Royston}.}
  \bibinfo{year}{1982}\natexlab{}.
\newblock \showarticletitle{Expected normal order statistics (exact and
  approximate)}.
\newblock \bibinfo{journal}{\emph{Journal of the Royal Statistical Society
  Series C (Applied Statistics)}} \bibinfo{volume}{31}, \bibinfo{number}{2}
  (\bibinfo{year}{1982}), \bibinfo{pages}{161--165}.
\newblock


\bibitem[Tang and Wang(2018)]%
        {tang2018personalized}
\bibfield{author}{\bibinfo{person}{Jiaxi Tang} {and} \bibinfo{person}{Ke
  Wang}.} \bibinfo{year}{2018}\natexlab{}.
\newblock \showarticletitle{Personalized Top-N Sequential Recommendation via
  Convolutional Sequence Embedding}. In \bibinfo{booktitle}{\emph{{WSDM} 2018,
  Marina Del Rey, CA, USA, February 5-9, 2018}}. \bibinfo{publisher}{{ACM}},
  \bibinfo{pages}{565--573}.
\newblock
\urldef\tempurl%
\url{https://doi.org/10.1145/3159652.3159656}
\showDOI{\tempurl}


\bibitem[Wang et~al\mbox{.}(2020)]%
        {wang2020off}
\bibfield{author}{\bibinfo{person}{Chengwei Wang}, \bibinfo{person}{Tengfei
  Zhou}, \bibinfo{person}{Chen Chen}, \bibinfo{person}{Tianlei Hu}, {and}
  \bibinfo{person}{Gang Chen}.} \bibinfo{year}{2020}\natexlab{}.
\newblock \showarticletitle{Off-Policy Recommendation System Without
  Exploration}. In \bibinfo{booktitle}{\emph{{PAKDD} 2020, Singapore, May
  11-14, 2020, Proceedings, Part {I}}} \emph{(\bibinfo{series}{Lecture Notes in
  Computer Science}, Vol.~\bibinfo{volume}{12084})}.
  \bibinfo{publisher}{Springer}, \bibinfo{pages}{16--27}.
\newblock
\urldef\tempurl%
\url{https://doi.org/10.1007/978-3-030-47426-3\_2}
\showDOI{\tempurl}


\bibitem[Wu et~al\mbox{.}(2019)]%
        {wu2019behavior}
\bibfield{author}{\bibinfo{person}{Yifan Wu}, \bibinfo{person}{George Tucker},
  {and} \bibinfo{person}{Ofir Nachum}.} \bibinfo{year}{2019}\natexlab{}.
\newblock \bibinfo{title}{Behavior Regularized Offline Reinforcement Learning}.
\newblock
\newblock
\showeprint[arxiv]{1911.11361}
\urldef\tempurl%
\url{http://arxiv.org/abs/1911.11361}
\showURL{%
\tempurl}


\bibitem[Wu et~al\mbox{.}(2021)]%
        {yuewu2021}
\bibfield{author}{\bibinfo{person}{Yue Wu}, \bibinfo{person}{Shuangfei Zhai},
  \bibinfo{person}{Nitish Srivastava}, \bibinfo{person}{Joshua~M. Susskind},
  \bibinfo{person}{Jian Zhang}, \bibinfo{person}{Ruslan Salakhutdinov}, {and}
  \bibinfo{person}{Hanlin Goh}.} \bibinfo{year}{2021}\natexlab{}.
\newblock \showarticletitle{Uncertainty Weighted Actor-Critic for Offline
  Reinforcement Learning}. In \bibinfo{booktitle}{\emph{{ICML} 2021, 18-24 July
  2021, Virtual Event}} \emph{(\bibinfo{series}{Proceedings of Machine Learning
  Research}, Vol.~\bibinfo{volume}{139})}. \bibinfo{publisher}{{PMLR}},
  \bibinfo{pages}{11319--11328}.
\newblock
\urldef\tempurl%
\url{http://proceedings.mlr.press/v139/wu21i.html}
\showURL{%
\tempurl}


\bibitem[Xin et~al\mbox{.}(2020)]%
        {xin2020self}
\bibfield{author}{\bibinfo{person}{Xin Xin}, \bibinfo{person}{Alexandros
  Karatzoglou}, \bibinfo{person}{Ioannis Arapakis}, {and}
  \bibinfo{person}{Joemon~M. Jose}.} \bibinfo{year}{2020}\natexlab{}.
\newblock \showarticletitle{Self-Supervised Reinforcement Learning for
  Recommender Systems}. In \bibinfo{booktitle}{\emph{{SIGIR} 2020, Virtual
  Event, China, July 25-30, 2020}}. \bibinfo{publisher}{{ACM}},
  \bibinfo{pages}{931--940}.
\newblock
\urldef\tempurl%
\url{https://doi.org/10.1145/3397271.3401147}
\showDOI{\tempurl}


\bibitem[Ye et~al\mbox{.}(2020)]%
        {ye2020mastering}
\bibfield{author}{\bibinfo{person}{Deheng Ye}, \bibinfo{person}{Zhao Liu},
  \bibinfo{person}{Mingfei Sun}, \bibinfo{person}{Bei Shi},
  \bibinfo{person}{Peilin Zhao}, \bibinfo{person}{Hao Wu},
  \bibinfo{person}{Hongsheng Yu}, \bibinfo{person}{Shaojie Yang},
  \bibinfo{person}{Xipeng Wu}, \bibinfo{person}{Qingwei Guo},
  \bibinfo{person}{Qiaobo Chen}, \bibinfo{person}{Yinyuting Yin},
  \bibinfo{person}{Hao Zhang}, \bibinfo{person}{Tengfei Shi},
  \bibinfo{person}{Liang Wang}, \bibinfo{person}{Qiang Fu},
  \bibinfo{person}{Wei Yang}, {and} \bibinfo{person}{Lanxiao Huang}.}
  \bibinfo{year}{2020}\natexlab{}.
\newblock \showarticletitle{Mastering Complex Control in {MOBA} Games with Deep
  Reinforcement Learning}. In \bibinfo{booktitle}{\emph{{AAAI}, New York, NY,
  USA, February 7-12, 2020}}. \bibinfo{publisher}{{AAAI} Press},
  \bibinfo{pages}{6672--6679}.
\newblock
\urldef\tempurl%
\url{https://ojs.aaai.org/index.php/AAAI/article/view/6144}
\showURL{%
\tempurl}


\bibitem[Yuan et~al\mbox{.}(2019)]%
        {yuan2019simple}
\bibfield{author}{\bibinfo{person}{Fajie Yuan}, \bibinfo{person}{Alexandros
  Karatzoglou}, \bibinfo{person}{Ioannis Arapakis}, \bibinfo{person}{Joemon~M.
  Jose}, {and} \bibinfo{person}{Xiangnan He}.} \bibinfo{year}{2019}\natexlab{}.
\newblock \showarticletitle{A Simple Convolutional Generative Network for Next
  Item Recommendation}. In \bibinfo{booktitle}{\emph{{WSDM} 2019, Melbourne,
  VIC, Australia, February 11-15, 2019}}. \bibinfo{publisher}{{ACM}},
  \bibinfo{pages}{582--590}.
\newblock
\urldef\tempurl%
\url{https://doi.org/10.1145/3289600.3290975}
\showDOI{\tempurl}


\bibitem[Zhang et~al\mbox{.}(2022)]%
        {zhang2020cost}
\bibfield{author}{\bibinfo{person}{Yifan Zhang}, \bibinfo{person}{Peilin Zhao},
  \bibinfo{person}{Qingyao Wu}, \bibinfo{person}{Bin Li},
  \bibinfo{person}{Junzhou Huang}, {and} \bibinfo{person}{Mingkui Tan}.}
  \bibinfo{year}{2022}\natexlab{}.
\newblock \showarticletitle{Cost-Sensitive Portfolio Selection via Deep
  Reinforcement Learning}.
\newblock \bibinfo{journal}{\emph{{IEEE} TKDE.}} \bibinfo{volume}{34},
  \bibinfo{number}{1} (\bibinfo{year}{2022}), \bibinfo{pages}{236--248}.
\newblock
\urldef\tempurl%
\url{https://doi.org/10.1109/TKDE.2020.2979700}
\showDOI{\tempurl}


\bibitem[Zhao et~al\mbox{.}(2019)]%
        {zhao2019deep}
\bibfield{author}{\bibinfo{person}{Xiangyu Zhao}, \bibinfo{person}{Changsheng
  Gu}, \bibinfo{person}{Haoshenglun Zhang}, \bibinfo{person}{Xiaobing Liu},
  \bibinfo{person}{Xiwang Yang}, {and} \bibinfo{person}{Jiliang Tang}.}
  \bibinfo{year}{2019}\natexlab{}.
\newblock \bibinfo{title}{Deep Reinforcement Learning for Online Advertising in
  Recommender Systems}.
\newblock
\newblock
\showeprint[arxiv]{1909.03602}
\urldef\tempurl%
\url{http://arxiv.org/abs/1909.03602}
\showURL{%
\tempurl}


\bibitem[Zhao et~al\mbox{.}(2018a)]%
        {zhao2018deep}
\bibfield{author}{\bibinfo{person}{Xiangyu Zhao}, \bibinfo{person}{Long Xia},
  \bibinfo{person}{Liang Zhang}, \bibinfo{person}{Zhuoye Ding},
  \bibinfo{person}{Dawei Yin}, {and} \bibinfo{person}{Jiliang Tang}.}
  \bibinfo{year}{2018}\natexlab{a}.
\newblock \showarticletitle{Deep Reinforcement Learning for Page-wise
  Recommendations}. In \bibinfo{booktitle}{\emph{{RecSys} 2018, Vancouver, BC,
  Canada, October 2-7, 2018}}. \bibinfo{publisher}{{ACM}},
  \bibinfo{pages}{95--103}.
\newblock
\urldef\tempurl%
\url{https://doi.org/10.1145/3240323.3240374}
\showDOI{\tempurl}


\bibitem[Zhao et~al\mbox{.}(2018b)]%
        {zhao2018recommendations}
\bibfield{author}{\bibinfo{person}{Xiangyu Zhao}, \bibinfo{person}{Liang
  Zhang}, \bibinfo{person}{Zhuoye Ding}, \bibinfo{person}{Long Xia},
  \bibinfo{person}{Jiliang Tang}, {and} \bibinfo{person}{Dawei Yin}.}
  \bibinfo{year}{2018}\natexlab{b}.
\newblock \showarticletitle{Recommendations with Negative Feedback via Pairwise
  Deep Reinforcement Learning}. In \bibinfo{booktitle}{\emph{{KDD} 2018,
  London, UK, August 19-23, 2018}}. \bibinfo{publisher}{{ACM}},
  \bibinfo{pages}{1040--1048}.
\newblock
\urldef\tempurl%
\url{https://doi.org/10.1145/3219819.3219886}
\showDOI{\tempurl}


\bibitem[Zheng et~al\mbox{.}(2018)]%
        {zheng2018drn}
\bibfield{author}{\bibinfo{person}{Guanjie Zheng}, \bibinfo{person}{Fuzheng
  Zhang}, \bibinfo{person}{Zihan Zheng}, \bibinfo{person}{Yang Xiang},
  \bibinfo{person}{Nicholas~Jing Yuan}, \bibinfo{person}{Xing Xie}, {and}
  \bibinfo{person}{Zhenhui Li}.} \bibinfo{year}{2018}\natexlab{}.
\newblock \showarticletitle{{DRN:} {A} Deep Reinforcement Learning Framework
  for News Recommendation}. In \bibinfo{booktitle}{\emph{{WWW} 2018, Lyon,
  France, April 23-27, 2018}}. \bibinfo{publisher}{{ACM}},
  \bibinfo{pages}{167--176}.
\newblock
\urldef\tempurl%
\url{https://doi.org/10.1145/3178876.3185994}
\showDOI{\tempurl}


\end{thebibliography}
\end{document}